\documentclass[10pt,journal,compsoc]{IEEEtran}

\ifCLASSOPTIONcompsoc
  \usepackage[nocompress]{cite}
\else
  \usepackage{cite}
\fi

\usepackage{graphicx}
\usepackage{subfigure}  

\usepackage{color}
\usepackage{multirow}
\usepackage{amsmath}
\usepackage{booktabs}
\usepackage{float}
\usepackage{bm}
\usepackage{wrapfig}
\usepackage{mathrsfs}
\usepackage{bbding}
\usepackage{amssymb}

\ifCLASSINFOpdf
\else
\fi

\begin{document}
%
\title{Improving Video Violence Recognition with Human Interaction Learning on 3D Skeleton Point Clouds}

\author{Yukun Su,
        Guosheng Lin,
        and~Qingyao Wu
\IEEEcompsocitemizethanks{\IEEEcompsocthanksitem Y. Su and Q. Wu are with the School of Software Engineering, South China University
	of Technology, Guangzhou 510640, China.\protect\\
E-mail: suyukun666@gmail.com, qyw@scut.edu.cn.
\IEEEcompsocthanksitem G. Lin is with the School of Computer Science and Engineering, Nanyang Technological University, Singapore. \protect\\
E-mail: gslin@ntu.edu.sg.
\IEEEcompsocthanksitem G. Lin and Q. Wu are the corresponding author.}}

\markboth{Journal of \LaTeX\ Class Files,~Vol.~14, No.~8, June~2021}%
{Shell \MakeLowercase{\textit{et al.}}: Bare Advanced Demo of IEEEtran.cls for IEEE Computer Society Journals}
%

\IEEEtitleabstractindextext{%
\begin{abstract}
Deep learning has proved to be very effective in video action recognition. Video violence recognition attempts to learn the human multi-dynamic behaviors in more complex scenarios. In this work, we develop a method for video violence recognition from a new perspective of skeleton points. Unlike the previous works, we first formulate 3D skeleton point clouds from human skeleton sequences extracted from videos and then perform interaction learning on these 3D skeleton point clouds. Specifically, we propose two types of Skeleton Points Interaction Learning (SPIL) strategies: (i) Local-SPIL: by constructing a specific weight distribution strategy between local regional points, Local-SPIL aims to selectively focus on the most relevant parts of them based on their features and spatial-temporal position information. In order to capture diverse types of relation information, a multi-head mechanism is designed to aggregate different features from independent heads to jointly handle different types of relationships between points. (ii) Global-SPIL: to better learn and refine the features of the unordered and unstructured skeleton points, Global-SPIL employs the self-attention layer that operates directly on the sampled points, which can help to make the output more permutation-invariant and well-suited for our task.
Extensive experimental results validate the effectiveness of our approach and show that our model outperforms the existing networks and achieves new state-of-the-art performance on video violence datasets.
\end{abstract}

}

\maketitle

\IEEEdisplaynontitleabstractindextext

%
\IEEEpeerreviewmaketitle

\ifCLASSOPTIONcompsoc
\IEEEraisesectionheading{\section{Introduction}\label{sec:introduction}}
\else
\section{Introduction}
\label{sec:introduction}
\fi

%
%
%
%
\IEEEPARstart{V}{ideo}-based violence recognition is defined as detecting violent behaviors in video data, which plays a crucial part in some video surveillance scenarios like railway stations, prisons or psychiatric centers. Different from the traditional video action recognition task, video violence recognition is more challenging because the video contains multiple characters and multi-dynamic features. 
	
\begin{figure}
	\begin{center}
		\centering
		\includegraphics[width=3.3in]{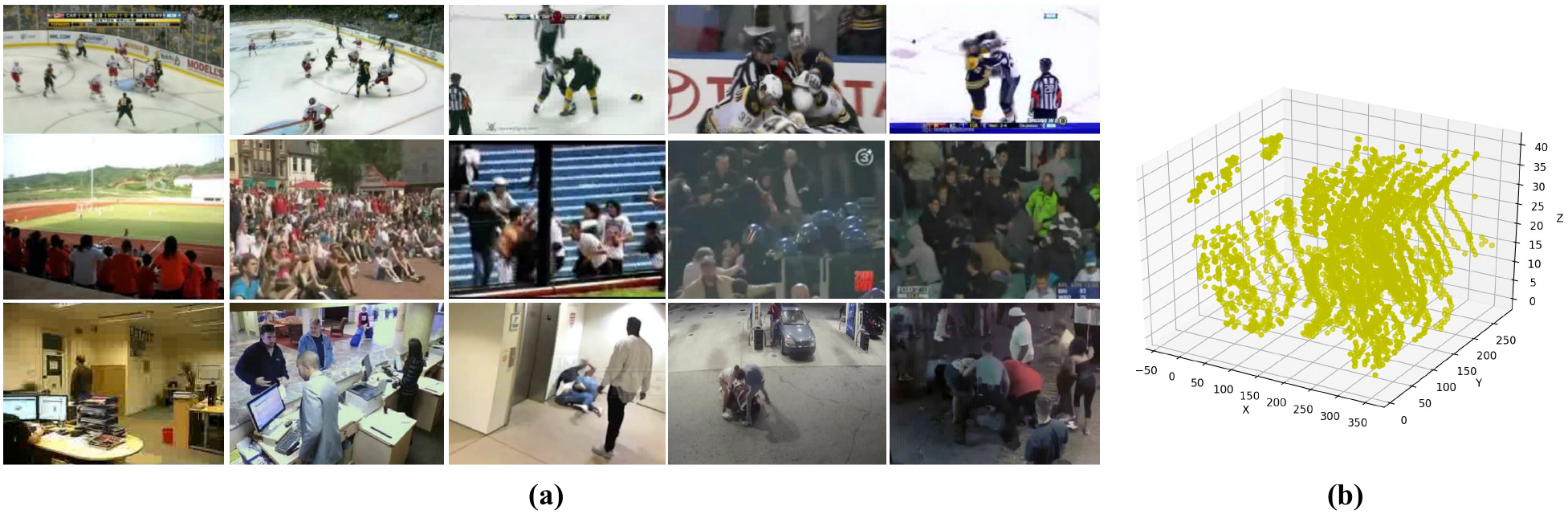}
	\end{center}
	\caption{(a) Some sample frames from the Hockey-Fight~\cite{hockey} dataset (1$^{st}$ row), the Crowd Violence~\cite{crowd} dataset (2$^{nd}$ row) and the RWF-2000 Violence~\cite{rwf} dataset (3$^{th}$ row). In each row, the left two columns are non-violent scenes while the right three columns are violent scenes. (b) 3D skeleton point clouds for a certain video.}
	\label{fig1}
\end{figure}

\begin{figure}
	\begin{center}
		\centering
		\includegraphics[width=3.3in]{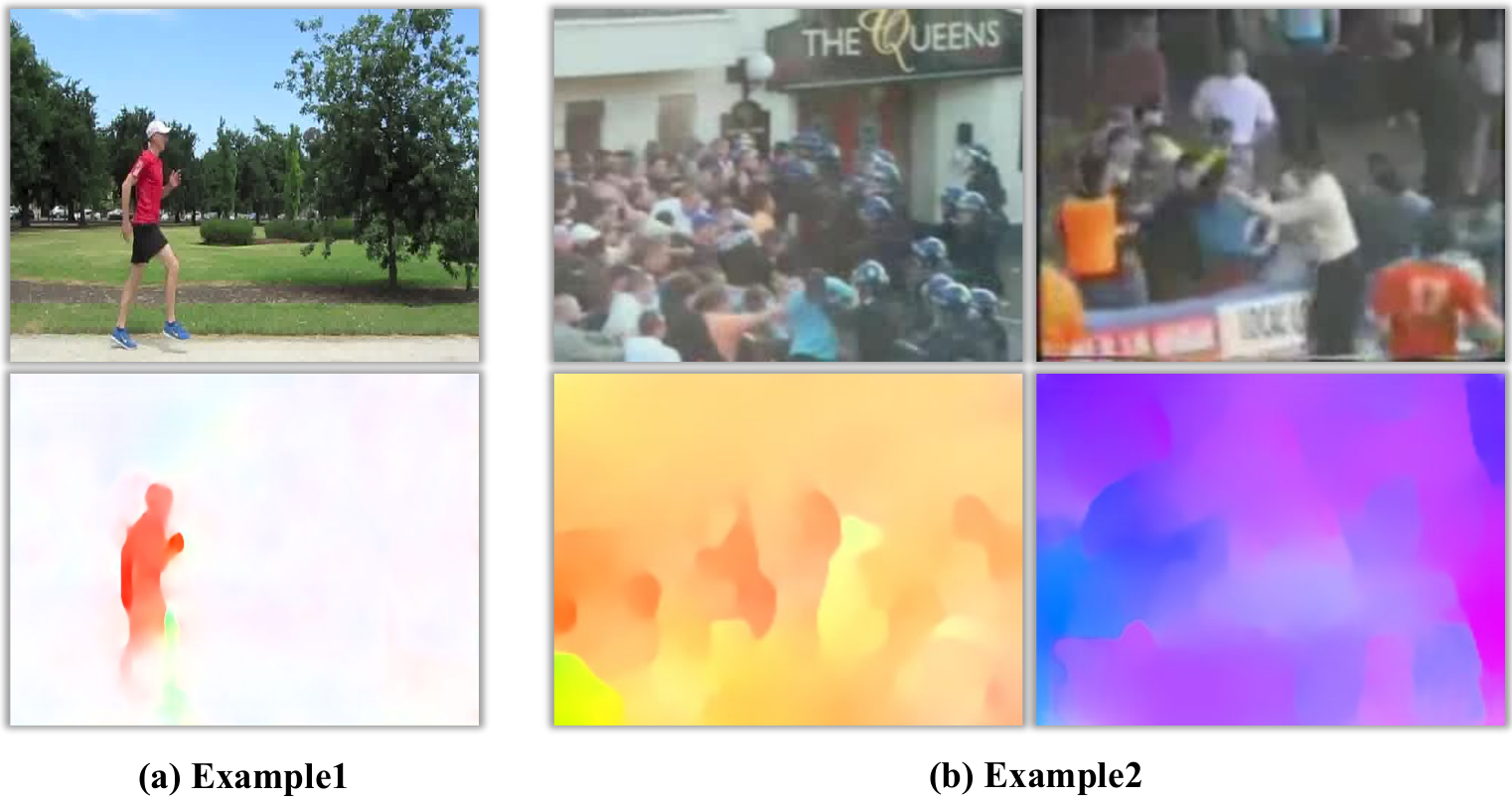}
		\includegraphics[width=3.3in]{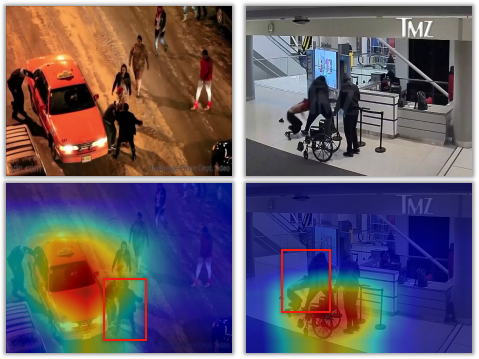}
	\end{center}
	\caption{Illustration of the failure cases of computing optical flow and activated maps in RGB-based method.}
	\label{fig666}
\end{figure}

Consider some sample frames from the public violence datasets, as shown in Fig~\ref{fig1}(a). When we humans see the sequences of images, we can easily recognize those violent actions through the human body's torso movements such as “kick", “beat", “push", etc.  Current deep learning approaches, however,  fail to capture these ingredients precisely in a multi-dynamic and complex multi-people scene. For example, the approaches based on two-stream ConvNets~\cite{two1,wang2016temporal} are learning to classify actions based on individual video frames or local motion vectors. However, such local motions that are captured by optical flow~\cite{brox2004high} sometimes fail to satisfy the dynamics modeling of shape change in multiple motion states. As shown in Fig~\ref{fig666} top, example 1 is captured from standard action recognition HMDB51~\cite{kuehne2011hmdb} dataset, where has only one person running. Therefore, the optical flow can be obtained perfectly. However, example 2 shows us some complex scenarios in violent videos. The optical flow fails to well reflect the object movements, which will yield unsatisfactory results in video violence recognition.

To tackle this limitation, recent Recurrent Neural or Vit~\cite{dosovitskiy2020image,su2023unified}Networks~\cite{donahue2015long,yue2015beyond} and 3D Convolutions~\cite{carreira2017quo,tran2015learning,tran2018closer} works have also focused on modeling long term temporal information.  However, all these frameworks focus on the features extracted from the whole scenes, leading to the interference by irrelevant information in the scenarios, and fail to capture region-based relationships as depicted in Fig~\ref{fig666} bottom. Meanwhile, the existing vision-based methods are mainly based on hand-crafted features such as statistic features between motion regions, leading to poor adaptability to another dataset. In violence recognition, extracting such appearance features and dynamics information of objects suffer from a number of complexities. Therefore, the above methods are often not very effective.

Inspired by the human vision, the movements of people in the video are reflected in human skeletal point sequences, which can be converted into 3D point clouds. We can then perform feature extraction on this data. Our experiments show that the existing 3D point clouds methods~\cite{qi2017pointnet++,wang2019dynamic,pointconv} can readily be applied to the violence recognition task. However, current methods, while excellent at extracting pertinent features on ordinary point clouds, they lack the ability to focus on relevant points and their interactions. 

To this end,  we propose a novel approach to perform video violence recognition via a human skeleton point convolutional reasoning framework. We first represent the input video as the cluster of 3D point clouds data as shown in Fig~\ref{fig1}(b) through extracting the human skeleton sequences pose coordinates from each frame in the video.
In order to better observe the characteristics of the skeleton points and conduct violence recognition, we introduce two types of Skeleton Points Interaction Learning (SPIL) strategies. Specifically, (i) Local-SPIL: the weight distribution among regional points with high coupling degree or strong semantic correlation is relatively high. Based on this, we can capture the appearance features and spatial-temporal structured position relation of skeleton points uniformly avoiding feature contamination between objects, and model how the state of the same object changes and the dependencies between different objects in frames. Besides, multiple heads attend to fuse features  from different independent heads to capture different types of information among points parallelly, which can enhance the robustness of the network. (ii) Global-SPIL: to further learn and refine the features on the global unordered points, a self-attention layer is proposed to operate directly on these sampled points. The output feature of each point is related to all input features, making it capable of learning the global context. The operations of the self-attention are parallelizable and order-independent, which can help to deal with the irregular data such as skeleton point clouds.

To validate our method, we conduct extensive experiments on multiple datasets to demonstrate the effectiveness of our algorithm. Our main contributions are summarized as follows:
\begin{itemize}
	\item 
	We formulate the video violence recognition task as 3D skeleton point clouds recognition problem, and we propose an effective interaction learning method on skeleton point clouds for video recognition.
	\item
	We propose a Local-SPIL to learn on the regional human skeleton points to capture both feature and position relation information simultaneously. And the multi-head mechanism allows to capture different types of points interactions to improve the network robustness.
	\item
	We propose a Global-SPIL to further extract and refine the features of the unordered and unstructured skeleton points, which enables the output more permutation-invariant and well-suited for our task.  
	\item 
	Different from the previous methods, we use the skeleton point clouds technique on recognition in violent videos, and our approach significantly outperforms state-of-the-art methods by a large margin. 
\end{itemize}

Our preliminary work was published in~\cite{su2020human}. To facilitate future research, the source code and trained models will be released publicly.

\section{Related Work}

\textbf{Video classification with deep learning}: Most recent works on video classification are based on deep learning. Initial approaches explored methods to combine temporal information based on pooling or temporal convolution~\cite{karpathy2014large,yue2015beyond}. To jointly explore spatial and temporal information of videos, 3D convolutional networks have been widely used.
Tran et al.~\cite{tran2015learning} trained 3D ConvNets on the large-scale video datasets, where they experimentally tried to learn both appearance and motion features with 3D convolution operations. In a later work, Hara et al.~\cite{hara2018can} studied the use of a ResNet~\cite{resnet} architecture with 3D convolutions and Xie et al.~\cite{xie2017aggregated} exploited aggregated residual transformations to show the improvements. Two-stream networks~\cite{carreira2017quo,i3d,wang2016temporal} have also been attracting high attention, they took the input of a single RGB frame (captures appearance information) and a stack of optical flow frames (captures motion information). An alternative way to model the temporal relation between
frames is by using recurrent networks~\cite{lev2016rnn,christoph2016spatiotemporal}.
However, these above approaches encountered bottlenecks in feature extraction when faced with more complex scenes and more irregular dynamic features in video violence recognition tasks. They fail to fully capture the comprehensive information in the entire video and are difficult to focus on distinguishing violent behavior in multiple characters and action features.

\begin{figure*}
	\begin{center}
		\centering
		\includegraphics[width=7.0in]{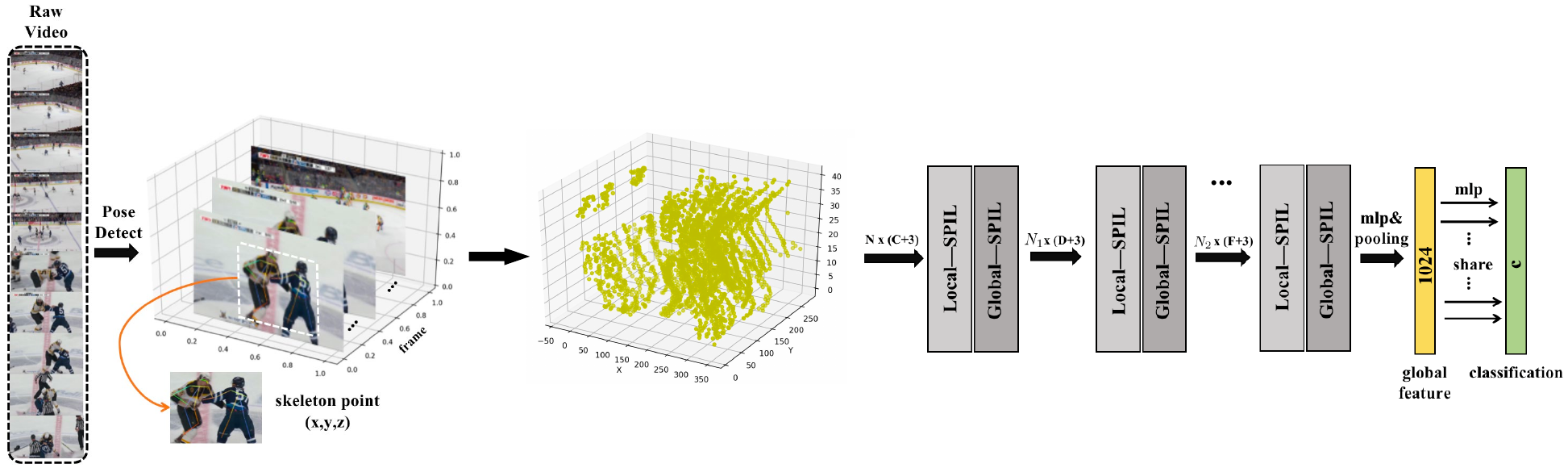}
	\end{center}
	\caption{The overview of our framework. Our model uses the pose detection method to extract the skeleton coordinates from each frame of the video. These human skeleton points are provided as point clouds inputs to  both Local-SPIL and Global-SPIL modules which perform information propagation based on assigning different weights to different skeleton points. Finally, a global pooling feature is extracted to perform classification.}
	\label{fig2}
\end{figure*}

\textbf{3D Point Clouds}: An alternative to represent the 3D human skeleton is 3D point clouds, which can model the human pose~\cite{su2022general} in high-level embedding.
To adapt the 3D points coordinates data for convolution, one straightforward approach is to voxelize it in a 3D grid structure~\cite{garcia2016pointnet,song2017semantic,su2021self,su2021modeling}. OctNet~\cite{riegler2017octnet} explored the sparsity of voxel data and alleviated this problem. 
However, since voxels are the discrete representations of space, this method still requires high-resolution grids with large memory consumption as a trade-off to keep a level of representation quality. 
Because the body keypoints themselves are a very sparse spatial structure, the application of the 3D voxel method to the skeleton points will lead to insufficient data characterization and make it difficult to train the model. 
In this trend, PointNet~\cite{qi2017pointnet} first discussed the irregular format and permutation invariance of
point sets, and presents a network that directly consumes
point clouds. PointNet++~\cite{qi2017pointnet++} extended PointNet by further considering not only the global information but also the local details with a farthest-sampling-layer and a grouping-layer. 
Deep learning in graph~\cite{bronstein2017geometric} is a modern term for a set of emerging technologies that attempts to address non-Euclidean structured data (e.g., 3D point clouds, social networks or genetic networks) by deep neural networks. Graph CNNs~\cite{niepert2016learning,defferrard2016convolutional} show advantages of graph representation in many tasks for non-Euclidean data, as it can naturally deal with these irregular structures.  ~\cite{zhang2018graph} built a graph CNN architecture to capture the local structure and classify point clouds, which also proves that deep geometric learning has enormous potential for unordered point clouds analysis.
Nonetheless, these works ignore the different importance of each point's contribution, especially in skeleton points. Even though some works~\cite{velivckovic2017graph,gehring2016convolutional,hao2023contrastive,cao4332138mutex} suggest the use of attention,  they seem to be of little use in the processing of specific spatial-temporal relation in the skeleton point clouds.

Different from these works, our approach encodes dependencies between objects with both feature and position relations, which focus on specific human dynamic action expressions ignoring action-independent information. This framework provides a significant boost over the state-of-the-arts.

\section{Proposed Method}

Our goal task is to represent the video as human skeleton point clouds of objects and perform reasoning for  video violence recognition. To this end, we propose the local and global Skeleton Points Interaction Learning (SPIL) modules to deal with the interrelationships between points. In this section, we will give detailed descriptions of our approach.~\ref{sec:frame} presents an overview of our framework.~\ref{sec:local} and~\ref{sec:global} introduce the detail of the Local- and Global-SPIL modules and layer updating, respectively.

\begin{figure}
	\begin{center}
		\centering
		\includegraphics[width=3.0in]{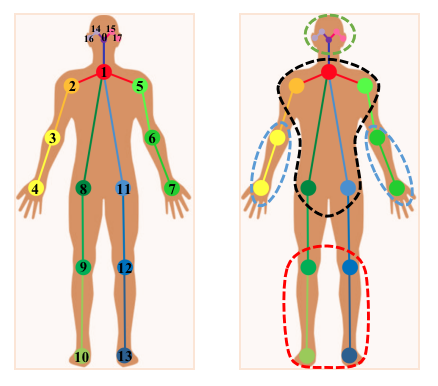}
	\end{center}
	\caption{The left sketch shows the joint label of the human kinetics-skeleton. Right: Green dot denotes the part of head.  Blue dot denotes the part of hands.  Red dot denotes the part of feet and the black dot denotes the part of central body.}
	\label{fig3}
\end{figure}

\begin{figure*}
	\begin{center}
		\centering
		\includegraphics[width=5.4in]{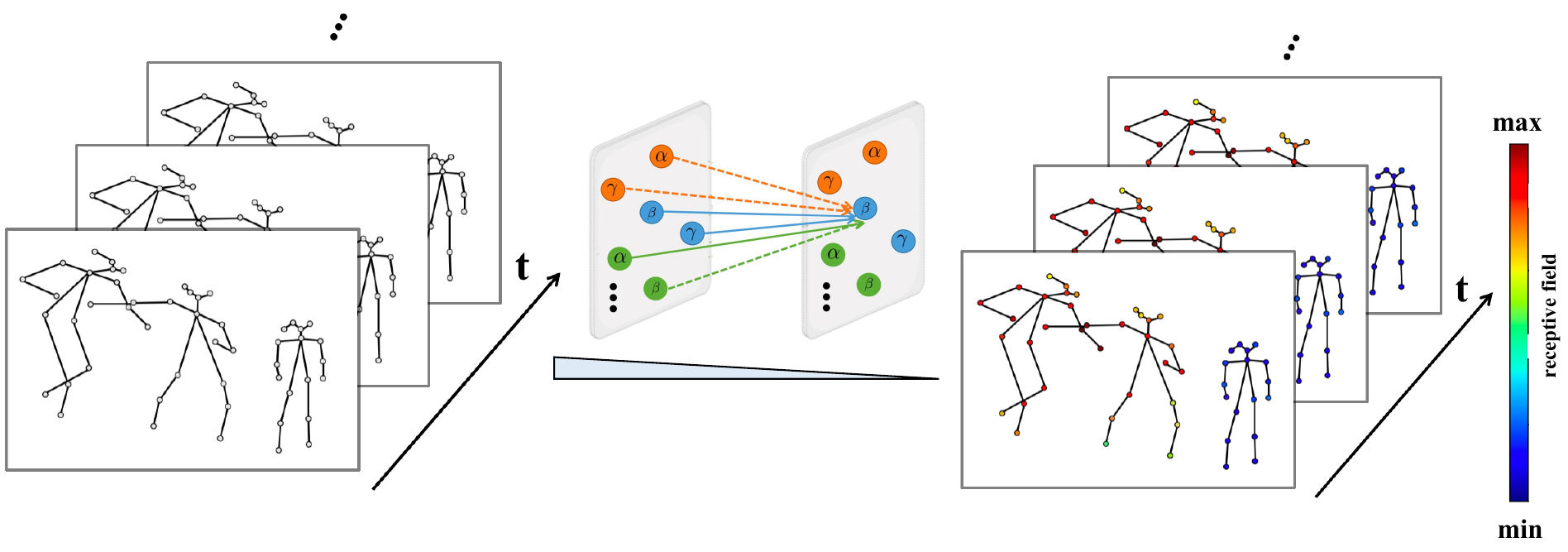}
	\end{center}
	\caption{Illustration of the Local-SPIL module on a subgraph of the human skeleton point clouds. Input points with a constant scalar feature (in grey) are convolved through a human skeleton points interaction filter. The output is a dynamically attentional weighted combination of the neighbor's points. The weights on irrelevant points  (the dotted arrows) are masked so that the convolution kernel can focus on the correlated points for prediction. ($\alpha$,$\beta$,$\gamma$ of different colors denote parts of different human joint points respectively, such as left hand, right elbow, left ankle).}
	\label{fig4}
\end{figure*}

\subsection{Framework}\label{sec:frame}

We propose to tackle this problem with an architecture, as illustrated in Fig~\ref{fig2}. Our approach takes raw video clips as input streams. First, we follow the human pose extraction strategy used in~\cite{fang2017rmpe}, which can detect body points from each frame in the video.  The coordinate ($x, y, z$) of each keypoint represents the position of the current point in each frame, where $x$ and $y$ represent the world coordinates of skeleton point of each frame and $z$ represents frame time $t$. Then we collect all the skeleton points sequences and transform the dynamic representation of the people in the video into a point clouds structure.

Subsequently, following the  scheme of~\cite{qi2017pointnet++}, we adopt the farthest point
sampling (FPS) algorithm to sample $N$ centroid points and take a $N\times (3+ C)$ matrix with 3-dim coordinates and $C$-dim points initial feature as input. For each centroid point with its $K$ neighbors points, the Local-SPIL abstraction level module is responsible for capture the spatial-temporal features in local region between the points. The Global-SPIL abstraction level module follows the Local-SPIL module which is responsible for learning and refining the $N$ global point features. They output a $N_1 \times (3 + D)$ matrix of $N_1$ subsampled points with 3-dim coordinates and new $D$-dim feature vectors representing updated features.
With this representation, we apply several both Local- and Global-SPIL modules to sample and extract the skeleton point clouds. Finally, the network classifies the global feature through a fully connected layer to conduct the violence video recognition.

Different from the~\cite{su2020human} that takes points’ confidence as initial feature. We consider that the contribution of different joint parts of the human body to the movement is inconsistent.
Movements such as "punching" and "kicking", which are involved in acts of violence, have more to do with the hands and feet. Therefore, in this work we assign different initial features to each joint depending on the part of the skeleton. As shown in Fig~\ref{fig3}, based on the joint label of the human skeleton, we separate the human skeleton into four parts: the head, the hands, the feet, and the central body. According to the aforementioned order of different parts, we assign them four different constants $\lambda_1$, $\lambda_2$, $\lambda_3$ and $\lambda_4$ that concatenate with their joints confidence together as initial features, respectively.

\begin{figure*}
	\begin{center}
		\centering
		\includegraphics[width=5.4in]{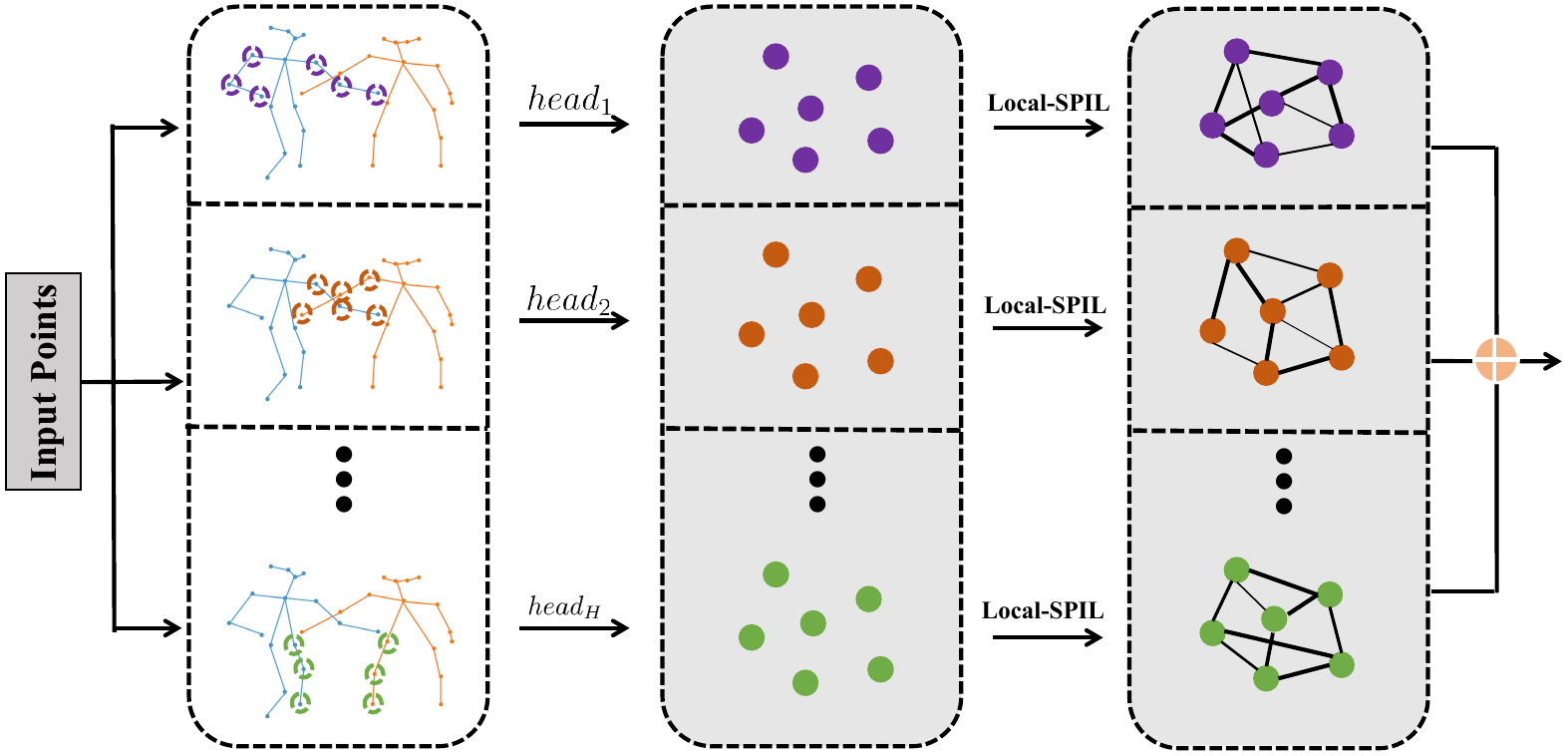}
	\end{center}
	\caption{Illustration of the multi-head Local-SPIL. A single head encodes the skeleton point clouds from the input independently, and multiple different headers are responsible for processing different types of information from the points and eventually aggregate them together. Each node denotes a skeleton joint point and each edge is a scalar weight, which is computed according to two points’ features and their relative position.}
	\label{fig5}
\end{figure*}

\subsection{Local-SPIL}\label{sec:local}

To conduct interaction learning and formulate the weight between regional points, the conventional approach is associated with the $K$ neighbors. However, not all nearby points have an effect on the current point. For instance, if there are multiple characters in a scene, irrelevant human skeleton points can sometimes be confusing and the learned feature characterizes all of its neighbors indistinguishably.
To address this problem, in our Local-SPIL module, as shown in Fig~\ref{fig4} intuitively, the points interaction weights on different skeleton points are distributed based on the relationships between points. We learn to mask or weaken part of the convolution weights according to the neighbors’ feature attributes. In this way, the network can focus on the skeleton points for prediction.

Given a point set $\left\{ p_1, p_2, ..., p_K \right\}$  $\in$ $\mathbb R^{3+C}$ according to a centroid point's $K$ neighbors, where the local region is grouped  within a radius. We set the radius to ($r \times T_{frame}$) that guarantees local region to cover more inter- and intra-frames points across space and time.
Particularly, the pair-wise interaction weights between points can be mathematically used $ W$ to represent, where the  weight $ W_{ij}$ indicates the connection of point $j$ to point $i$. In the case of $ W \in \mathbb R^{K\times K}$, we define the set of points as $\mathcal S$ = $\left\{ (p^f_i,p^l_i) \rvert i = 1,...,K\right\}$. Among them, $p^f_i \in \mathbb R^C$ is point $i$'s $C$-dim initial feature, and $p^l_i =(l^\text{x}_i,l^\text{y}_i,l^\text{z}_i) \in \mathbb R^3$ is the 3-dim position  coordinates.

Traditional graph attention NNs~\cite{defferrard2016convolutional,zhang2018graph} learn to capture the local structure, however, they only consider the feature term. Unlike the traditional point operation, for the distribution of the weights $W$, in order to obtain distinguishable representation ability to capture the correlation between different skeletal points, it is necessary to consider both feature similarity and position characteristic relation.
To this end, we separately explore the feature and position information and then perform high-level modeling of them to dynamically adapt to the structure of the objects. Concretely, the interaction weight of each neighboring point is computed as follows:

\begin{equation}
W_{ij} = \Phi( R^ F(p^f_i,p^f_j), R^ L(p^l_i,p^l_j)), \tag{1}
\label{eq1}
\end{equation}
where $ R^ F(p^f_i,p^f_j)$ implies the feature relation between points and $ R^ L(p^l_i,p^l_j)$ denotes the position relation. $\Phi$ function plays the role of a combination of feature information and position information.

In this work, we use the differentiable architecture as in~\cite{wu2019learning} but different in building $ R^ F$ and $ R^ L$ functions detailedly to compute points interaction value. Then, Eq(\ref{eq1}) can be reformulated as follows:

\begin{equation}
W_{ij}=\frac{ R^ L(p^l_i,p^l_j) \ exp(  R^F(p^f_i,p^f_j))}{\sum_{j=1}^{K} R^ L(p^l_i,p^l_j) \ exp( R^ F(p^f_i,p^f_j))}, \tag{2}
\label{eq2}
\end{equation}
where the points interaction weights are normalized across all the neighbors of a point $i$ to handle the size-varying neighbors across different points and spatial scales.  Formally, it is the weighted average of $R^ L(p^l_i,p^l_j)$, and the coefficient is $R^ F(p^f_i,p^f_j)$.

\textbf{Feature Term}: 
Intuitively, the points of different features in the local area exert various influences to enhance the expressive power of each point. Our feature modulator solves this problem by adaptively learning the amount of influence each feature in a point has on other points. We can utilize the dot-product operation to compute similarity in embedding space, and the corresponding function $ R^ F$can be expressed as:

\begin{equation}
{R^{F}}(p^f_i,p^f_j)= \phi(g(p^f_i))^\mathsf{T}\theta(g(p^f_j)),\tag{3}
\label{eq3}
\end{equation}
where $g$(·): $\mathbb R^C$ $\to$ $\mathbb R^{C'}$ is a feature mapping function. $\phi$(·) and $\theta$(·) are two learnable linear projection functions, followed by ReLU, which project the features relation value between two points to a new space.

\textbf{Position Term}: 
In order to make full use of the spatial-temporal structure relation of points, the position characteristics of points should be taken into account. In our work, we consider the following three choices:

(1) \ {Euclidean-Distance Spacing}: Considering the Euclidean distance between the points, the relatively distant points contribute less to the connection of the current point than the nearby points. With this in mind, we directly calculate the distance information and conduct on the points. The $ R^ P$ can be formulated as: 

\begin{equation}
R^L(p^l_i,p^l_j) = -ln(\sigma(\mathcal D(p^l_i,p^l_j))), \tag{4}
\label{eq4}
\end{equation}
where $\sigma$(·) is a sigmoid activation function and the output range is controlled between [0, 1] to feed to the $\mathit{ln}$ function. $\mathcal D$ is a function to calculate the distance between points.

(2)\ {Euclidean-Distance Spanning}: Alternatively, we can first encode the  relations between two points to a 
high-dimensional representation based on the position distance. Then the difference between the two terms encourages to span the relations to a new subspace. Specifically, the position relation value is computed as:

\begin{equation}
R^L(p^l_i,p^l_j) = \frac{\psi(M_1(p^l_i) - M_2(p^l_j))}{\mathcal D(p^l_i,p^l_j)}, \tag{5}
\label{eq5}
\end{equation}
where $M_1$(·) and $M_2$(·) are two multilayer perceptrons functions and $\psi$(·) is a linear projection function followed by ReLU that generates the embedded feature into a scalar.

(3)\ {Euclidean-Distance Masking}: In addition, a more intuitive approach is to ignore some distant points and retain the characteristic contribution of the relative nearby points. 
For this purpose, we set a threshold to ignore the contribution of certain points and  the function can be defined as:

\begin{equation}
\begin{split}
{R^{L}}(p^l_i,p^l_j)= \begin{cases}
\ \ \ \ \ \ \ \  \ \ \ \ \ 0  &, \ if \  \underset{(l^z_i = l^z_j)}{\mathcal D}(p^l_i,p^l_j) > d,  \\
\psi(M_1(p^l_i) || M_2(p^l_j))&, \ else. 
\end{cases} \end{split}  \tag{6}
\label{eq6}
\end{equation}

Note that the radius ($r \times T_{frame}$) ensures the local region cover points spatially and temporally. However, spatially, we try to mask out some weak correlation points within the same frames. The implication is that we preserve the globality in time and the locality in space. $||$ is the concatenation operation and the embedded feature between two points is transformed into a scalar by a learnable linear function, followed by a ReLU activation. $d$ acts as a distance threshold which is a hyper-parameter.

\textbf{Multi-head mechanism}: 
Although a single head Local-SPIL module can perform interaction feature extraction on skeleton points, since the connection between the joint points of the human body is ever-changing, each point may have different types of features.
For example, a joint point has an information effect on its own posture and also a dynamic information effect on the interaction between human bodies at the same time, we term it a point with different types of features. Specifically, as shown in Fig~\ref{fig5}, for a certain skeletal point such as the elbow joint, the first head may be sensitive to the information of the human elbow joint's own posture, such as judging whether it is a “punch” posture; while the second head is more concerned with the connection of elbow joint motion information between people to extract dynamic features. In the same way, the remaining ($H$-2) heads extract different features for other types that may be related.

For this reason, the designed multi-head mechanism allows the Local-SPIL module to  work in parallel to capture diverse types of relation points. Every weight ${W}_\iota$ is computed in the same way according to Eq(\ref{eq2}), where $\iota$ $\in$ $H$ is the number of heads.
It should be noted that independent head does not share weights during the calculation.
By using the multi-head mechanism, the model can make more robust relational reasoning upon the points.

\textbf{Local layer updating}: 
To perform reasoning on the local skeleton points, unlike the standard 2D or 3D convolutions that run on a local regular grid. For a target point, the output are the updated features of each object points from all its neighbors. We can represent one layer of convolutions as:

\begin{equation}
X^{(l+1)} = {W}X^{(l)} \mathcal{M}^{(l)},\tag{7}
\label{eq7}
\end{equation}
where $ W$ $\in$ $\mathbb{R}^{K\times K}$ represents the interaction weights. $X^{(l)} \in \mathbb{R}^{K\times C'}$ is the input feature projected by $g$(·) mapping function of the centroid grouping skeleton point set. $\mathcal M^{(l)} \in \mathbb{R}^{C'\times d}$ is the layer-specific learnable weight matrix.
After each layer of convolutions, we adopt non-linear functions for activating before the feature $X^{(l+1)}$ is forwarded to the next layer. 

\begin{figure}
	\begin{center}
		\centering
		\includegraphics[width=3.3in]{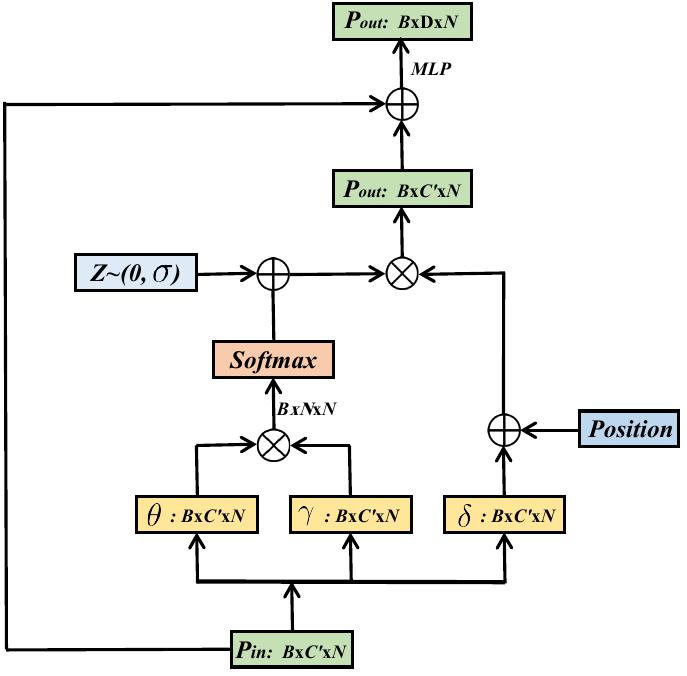}
	\end{center}
	\caption{Illustration of the Global-SPIL module layer. The output is associated with all input features and the self-attention operation are parallelizable and order-independent. $\bigotimes$ indicates the matrix multiplication and $\bigoplus$ indicates the matrix elementwise summation.}
	\label{fig6}
\end{figure}

\begin{figure}
	\begin{center}
		\centering
		\includegraphics[width=3.3in]{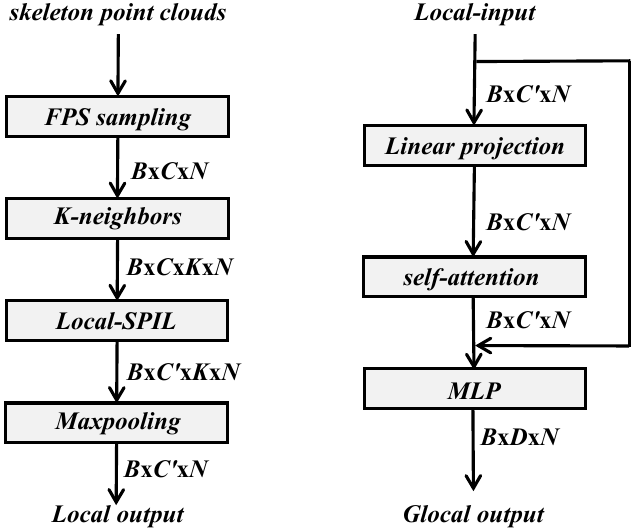}
	\end{center}
	\caption{ The detailed illustration of the designed Local- and Global-SPIL block structures. The input of the Global-SPIL module is the output of the Local-SPIL module.}
	\label{fig7}
\end{figure}

To combine multi-head weights, in this work, we employ the 
concatenation fusion function. We can extend Eq(\ref{eq7}) as:

\begin{equation}
X^{(l+1)} ={\overset{H} {\underset{\iota}{\|}}} ({W}_\iota X^{(l)} \mathcal M^{(l)}_\iota, dim =1), \tag{8}
\label{eq8}  
\end{equation}
where $ W_\iota$ indicates different types of weights and the different $\mathcal{M}_\iota$ are not shared. $\|$(·) function aggregates and fuses the output information of all $H$ heads. Namely, all $K \times d$-dim features will concatenate together to form the new features. Thus the points can be increased in dimensionality to obtain more characteristic information. Afterward, for each subsampled centroid grouping points, $Maxpooling$ is applied for fusing $K$ local region points that are updated by Eq(\ref{eq8}).

\subsection{Global-SPIL}\label{sec:global}

Since the 3D skeleton point clouds are unstructured unlike the 2D images, thus,  the sampled $N$ centroid  points are orderless. Although we have captured the relational features within the local points by $K$ neighbors in~\ref{sec:local}, it is equally crucial to process the point clouds learning features upon the global sampled points to make the output more stable. Inspired by the attention mechanism~\cite{attention} in natural language processing field, which is inherently permutation invariant for processing the sequence data. To better refine the features of the skeleton points, we propose to tackle the drawbacks by using a self-attention module layer, as shown in Fig~\ref{fig6}. 

Considered $\mathcal{X}$ = $\left\{ x_i \right\} _{N}$ be a set of global sampled points. The conventional standard attention operation in~\cite{attention} can be formulated as follows:

\begin{equation}
y_i = \sum_{x_j \in \mathcal{X}}^{} \text{softmax} (\theta(x_i) ^\mathit{T} \gamma(x_j) \delta(x_j)),\tag{9}
\label{eq9}
\end{equation}
where $\theta$(·), $\gamma$(·) and $\delta$(·) are three different learnable linear projection functions and  $y_i$
is the output feature.  The attention layer computes the scalar product between features transformed by $\theta$
and $\gamma$ and uses the output as an attention weight for aggregating features transformed by $\delta$.

However, directly applying the above attention mechanism to skeleton point clouds is trivial, since the skeleton points have a difference of position and the global centroid points are randomly sampled. Therefore, we modify Eq(\ref{eq9}) and propose our Global-SPIL self-attention layer to make the output more permutation-invariant and robust, which can be formulated as follows:

\begin{equation}
y_i = \sum_{x_j \in \mathcal{X}}^{} MLP(\text{softmax} ((\theta(x_i) ^\mathit{T} \gamma(x_j) + \mathcal{Z}) (\delta(x_j) + \xi)) + x_i),\tag{10}
\label{eq10}
\end{equation}
where MLP indicates the multilayer perceptions function. $\mathcal{Z}$ is an $B \times N \times$N matrix in $B$ batchsize and $N$ point number. The elements of $\mathcal{Z}$ are parameterized and optimized together with the other parameters in the training process. With this data-driven manner, the network can learn global context that are fully targeted to the recognition task and more individualized for different information contained in different points under the perturbation of random sampling. Additionally, we take skeleton position information into account to adapt to global structure in the data. Concretely, the position embedding term $\xi = \eta(x_i - x_j)$. Here, we utilize the subtraction relation and $x_i$, $x_j$ are the 3D point coordinates for points $i$ and $j$. The encoding function $\eta$(·) is two layers learnable linear projection function followed by ReLU for nonlinearity.
A residual connection is added for each Global-SPIL layer, which allows the layer to have better information propagation without dropping its original features.

The detailed structures of the proposed Local- and Global-SPIL modules are shown in Fig~\ref{fig7}. Given a skeleton point clouds sample, we first calculate the local interaction relations between regional points. Then, the output is fed into the global module for further feature learning. We stack several both SPIL blocks in our network. 
Finally, as illustrated in Fig~\ref{fig2}, the output relational features are forwarded to a global average pooling layer, which calculates the mean of all the proposal features and leads to a $1 \times 1024$ dimensions representation. Then it is fed to the classifier to generate predictions. ${Y}$ denotes the labels and $\hat{Y}$ are predictions, with standard cross-entropy loss, the final loss function is formed as:

\begin{equation}
\mathcal{L}oss = -Ylog\hat{Y} + (1-Y)log(1- \hat{Y}). \tag{11}
\label{eq11}  
\end{equation}

\section{Experiments}
In this section, first, we describe the datasets in experiments
and then provide implementation details. Second, we conduct ablation studies to explore the components of our proposed approach. Last, we compare
the proposed method with the state-of-the-art methods.

\begin{table}[]
	\begin{center}
		\caption{Exploration of different skeleton points interaction learning strategies for violence recognition.}\label{table1}
		\scalebox{0.9}{
			\begin{tabular}{cccccc}
				\toprule  
				\toprule  
				Baseline  & Spacing & Spanning & Masking & Global-SPIL & Acc.(\%) \\
				\midrule  
				$\checkmark$&  & & & & 84.3 \\
				$\checkmark$&  $\checkmark$ & &  & & 86.6 \\
				$\checkmark$&  & $\checkmark$ & &  & 88.2 \\
				$\checkmark$&  &  & $\checkmark$ &  & \textbf{88.8} \\
				\midrule  
				\midrule  
				$\checkmark$&  $\checkmark$ & &  & $\checkmark$& 87.0 \\
				$\checkmark$&  & $\checkmark$ & &  $\checkmark$& 88.6 \\
				$\checkmark$&  &  & $\checkmark$ &  $\checkmark$& \textbf{89.3} \\
				\bottomrule 
		\end{tabular}}
	\end{center}
\end{table}

\begin{table}
	\begin{center}
		\caption{Exploration of number of multi-heads.}
		\label{table2}
		\scalebox{0.8}{
			\begin{tabular}{c|c|c|c|c|c|c|c}
				\toprule
				Num-heads \ & \ $1$ \ & \ $2$ \ & \ $4$ \ & \ $8$ \ & \  $16$  \ & \ $32$ \ & \ $64$ \ \\
				\midrule
				LG-SPIL(Ours) (\%) \ & \ 89.3 \ & \ 89.5 \ & \ 89.4 \ & \ \textbf{89.8} \ & \ 89.6 \ & \ 89.5 \ & \ 89.6 \ \\
				\midrule
				Baseline (\%) \ & \ 84.3 \ & \ 84.6 \ & \ \textbf{84.8} \ & \ 84.7 \ & \ 84.7 \ & \ 84.6 \ & \ 84.6 \ \\
				\bottomrule
		\end{tabular}}
	\end{center}
\end{table}

\subsection{Datasets}

We train and evaluate our model on five datasets (Hockey-Fight dataset~\cite{hockey}, Crowd Violence dataset~\cite{crowd}, Movies-Fight dataset~\cite{movie}, RWF-200 Violence dataset~\cite{rwf} and the AVDV dataset~\cite{bianculli2020dataset}) for video violence recognition.
To the best of our knowledge, RWF-2000 is the largest dataset  which consists of 2,000 video clips captured by surveillance cameras in real-world scenes. Each video file is a 5-second video clip with 30 fps. Half of the videos contain violent behaviors, while others belong to non-violent actions. More specifically, Hockey-Fight, Crowd Violence, Movies-Fight and the AVDV datasets compose of 1000, 246, 200 and 350 clips, respectively. In datasets such as the Crowd Violence dataset and AVDV, they both have long-term videos with the longest 14s.
All videos in these datasets are captured by surveillance cameras in the real world, none of them are modified by multimedia technologies. 
Unlike the ordinary action video datasets, where only one or two people are existing in the scenes, these violent video datasets usually have a lot of characters in each video and they are not fixed. The background information is complicated and the dynamic information of each character is also more changeable.

\subsection{Implementation Details}

Our implementation is based on PyTorch deep learning framework. Under the setting of batch-size 8 and 2048 sampled points, dropout~\cite{dropout} is applied to the last global fully connected  layer with a ratio of 0.4. We train our model from scratch for 200 epochs using stochastic gradient descent (SGD) optimizer with momentum and a learning rate of 0.001 on 4 Nvidia 2080Ti GPUs. Empirically, we set the layer number of the Local- and Global-SPIL modules to 3 with  radius $r$ = 0.8, 0.6, 0.4 in turn and $T_{frame}$ = 5. $\lambda_1$, $\lambda_2$, $\lambda_3$ and $\lambda_4$ are set to 0.2, 0.8, 0.6 and 0.4, respectively. Augmentation is used during the training with randomly jittering and rotating($\pm$10\%). 
For the baseline of our skeleton point convolution network, we give a vanilla model without using SPIL and multi-head mechanism. Namely, we simply use the feature information of the points and single head, ignoring the position information, and then use the multi-layer perceptrons to perform operations.
We run all experiments four times with different random seeds and report mean accuracies.
For simplicity, we define the method using Eq(\ref{eq4}) as \textbf{Spacing}, and \textbf{Spanning} represents using Eq(\ref{eq5}) while \textbf{Masking} implies Eq(\ref{eq6})'s strategy.

\begin{table}
	\begin{center}
		\caption{ Exploration of different distance threshold $d$ in Eq~\ref{eq6}.}
		\label{table3}
		\scalebox{0.9}{
			\begin{tabular}{c|c|c|c|c}
				\toprule
				Euclidean-Distance \ &  $d_1 = 0.01$ \ &  $d_2 = 0.02$ \ &  $d_3 = 0.04$ \ &  $d_4= 0.08$ \  \\
				\midrule
				Acc.(\%) \ & \ 89.7 \ & \ 89.8 \ & \ \textbf{90.0} \ & \ 89.7 \ \\
				\bottomrule
		\end{tabular}}
	\end{center}
\end{table}

\begin{table}
	\begin{center}
		\caption{ Exploration of different initial features for network learning. $||$ indicates the concatenate operation.}
		\label{table4}
		\scalebox{0.85}{
			\begin{tabular}{c|c|c|c}
				\toprule
				Initial features \ &  confidence \ &   body parts \ &   confidence $||$ body parts (Ours)\\
				\midrule
				Acc.(\%) \ & \ 89.6 \ & \ 89.8 \ & \ \textbf{90.0} \  \\
				\bottomrule
		\end{tabular}}
	\end{center}
\end{table}

\subsection{Ablation Studies}

To explore the components of our proposed method, we first conduct extensive analysis on RWF-2000 datasets~\cite{rwf} to demonstrate how they help to improve feature learning for recognition. 

\vspace{1ex}

\noindent \textbf{The effect on Local- and Global-SPIL}: To show the impact of the Local- and Global-SPIL modules, we compare the baseline model with the proposed interaction learning modules we designed. Among them, we all use a single head method for a fair comparison. At the same time, we set the default value as 0.02 in the hyper-parameter of the threshold $d$ in Eq(\ref{eq6}).
As shown in table~\ref{table1}, in the case of point convolution operation without using SPIL to extract both feature and position information of points, our baseline model has an accuracy of only \textbf{84.3\%}. After adopting the proposed Local-SPIL module, base on a single head, all methods outperform the based model, demonstrating the effectiveness of modeling interaction weights between points. And the \textbf{Masking} yields the best accuracy with \textbf{88.8\%} than the other two ways. We conjecture that the function of the mask can filter out some redundant information and make the information more stable. Additionally, by jointly employing Local- and Global-SPIL strategy, we can further improve the performance. Specifically, \textbf{Masking} combining with the \textbf{Global-SPIL} strategy can boost the baseline performance achieving the accuracy with \textbf{89.3\%}, which validate the superiority of the proposed Global-SPIL module.
Note that, in the rest of the paper, we choose Masking \textbf{L}ocal-SPIL combining with the \textbf{G}lobal-SPIL strategy to represent our main method, termed as LG-SPIL.

\vspace{1ex}

\noindent \textbf{The effect on Multi-heads}: We also reveal the effectiveness of building a multi-head mechanism to capture diverse types of related information. As depicted in table~\ref{table2},  we compare the performance of using different numbers of heads in baseline and our method, which the results indicate that multi-head mechanism can improve both models in different degrees. When the head number is set to 8, our \textbf{LG-SPIL} is able to further boost accuracy from \textbf{89.3\%} to \textbf{89.8\%}. In contrast, too many heads will lead to redundant learning features and will increase the computational costs.

\begin{table}[]
	\begin{center}
		\caption{Exploration of different composition of the self-attention mechanism for violence recognition.}\label{table5}
		\scalebox{0.9}{
			\begin{tabular}{ccccc}
				\toprule  
				\toprule  
				Attention  & + $\mathcal{Z}$ & + Position term & + Residual connection & Acc.(\%) \\
				\midrule  
				$\checkmark$&  & & & 89.2 \\
				$\checkmark$&  $\checkmark$ & & & 89.5 \\
				$\checkmark$&  & $\checkmark$&  & 89.6 \\
				$\checkmark$&  $\checkmark$& $\checkmark$ &   & 89.8 \\
				$\checkmark$&  $\checkmark$& $\checkmark$ & $\checkmark$   & \textbf{90.0} \\
				\bottomrule 
		\end{tabular}}
	\end{center}
\end{table}

\begin{table}[]
	\begin{center}
		\caption{Comparison with state-of-the-arts on the RWF-2000 dataset.}\label{table6}
		\scalebox{1.0}{
			\begin{tabular}{ccc}
				\toprule  
				\toprule  
				Method \ &  \ Core Operator \ & \ Acc.(\%)\\
				\midrule  
				TSN~\cite{tsn} \ & \ Two-Stream \ & \ 81.5\\
				I3D~\cite{i3d} \ & \ Two-Stream \ & \ 83.4\\
				3D-ResNet101~\cite{hara2018can} \ & \ 3D Convolution \ & \ 82.6\\
				ECO~\cite{eco} \ & \ 3D Convolution + RGB \ & \ 83.7\\
				Representation Flow~\cite{piergiovanni2019representation} \ & \ Flow + Flow \ & \ 85.3\\
				Flow Gated Network~\cite{rwf} \ & \ 3D Convolution + flow \ & \ 87.3\\
				TEA~\cite{tea} \ & \ 2D Convolution \ & \ 86.9\\
				TIN~\cite{tin} \ & \ Temporal interlacement \ & \ 87.2\\
				\midrule  
				PointNet++~\cite{qi2017pointnet++} \ & \  Multiscale Point MLP \ & \ 78.2\\
				PointConv~\cite{pointconv} \ & \ Dynamic Filter \ & \ 76.8\\ 
				DGCNN~\cite{wang2019dynamic} \ &\  Graph Convolution \ & \ 80.6\\
				L-SPIL~\cite{su2020human} \ & \ {Local-SPIL } \ & \ {89.3}\\
				\textbf{LG-SPIL(Ours)} \ & \ \textbf{Local \& Gloabal SPIL } \ & \ \textbf{90.0}\\
				\bottomrule 
		\end{tabular}}
	\end{center}
\end{table}

\vspace{1ex}

\noindent \textbf{The effect on Hyper-parameter}: Furthermore, we also implement experiments to reveal the effect of the hyper-parameter threshold of $d$ on network performance. As shown in table~\ref{table3}, we can see that when $d$ = 0.04 achieves the best performance approaching \textbf{90.0\%}. We conjecture that a too-small threshold will cause network information to be lost, and a too-large threshold will cause a negative effect of excess network information. Thus, we adopt head number $H$= 8, $d$= 0.04 in the following experiments.

\vspace{1ex}

\noindent \textbf{The effect on Initial features}: As aforementioned, we divide the human body into four parts according to the movement attribute and assign them different initial values. As shown in table~\ref{table4}, the proposed different body parts values concatenating with corresponding joints confidence can get the best performance, which shows that rich initial features information can help the subsequent learning of the network.

\vspace{1ex}

\noindent \textbf{Exploration on Self-attention}: Since the self-attention~\cite{attention} can be directly applied in our method, we conduct experiments to show our proposed improved self-attention can be better  adapted to our task.
As shown in table~\ref{table5}, directly deploying attention in~\cite{attention} can only achieve the accuracy of 89.2\%. When we inject $\mathcal{Z}$ and the position term, respectively, our approach can boost the performance. 
It's worth noting that when the three components ($\mathcal{Z}$, position term and the residual connection) combining together, the network can further yield the best output. This reveals the effectiveness of our proposed global self-attention module and show that our method can fully adapt to the unstructured skeleton points data.

\subsection{Comparison with the State-of-the-Art}

\begin{table}[]
	\begin{center}
		\caption{Comparison with state-of-the-arts on the Hockey-Fight, Crowd and Movies-Fight dataset.}\label{table7}
		\scalebox{0.9}{
			\begin{tabular}{cccc}
				\toprule  
				\toprule  
				Method &  Hocky-Fight  & Crowd & Movies-Fight \\
				\midrule  
				3D CNN~\cite{ding2014violence}   & 91.0 & - & -  \\
				Extreme Acceleration~\cite{deniz2014fast}   & - &-& 85.4\\
				MoWLD + BoW~\cite{zhang2017mowld}  & 91.9 & 93.1& - \\
				TSN~\cite{tsn} &  91.5 &  81.5 & 94.2\\
				I3D~\cite{i3d}  & 93.4  &  83.4 & 95.8\\
				3D-ResNet101~\cite{hara2018can} &  93.6  &  84.8 & 96.9\\
				ECO~\cite{eco}  & 94.0  &  84.7 & 96.3\\
				Representation Flow~\cite{piergiovanni2019representation}  & 92.5 & 85.9 & 97.3\\
				Flow Gated Network~\cite{rwf}  & \textbf{98.0} & 88.8 & - \\
				TEA~\cite{tea}  &  97.1 &  89.5 & 98.0\\
				TIN~\cite{tin}  & 96.8  &  88.3 & 97.4\\
				\midrule  
				PointNet++~\cite{qi2017pointnet++}  & 89.7 &  89.2 & 89.2\\
				PointConv~\cite{pointconv}  & 88.6 & 88.9 & 91.3 \\ 
				DGCNN~\cite{wang2019dynamic}  & 90.2 & 87.4 & 92.6 \\
				L-SPIL~\cite{su2020human}  & 96.8 & 94.5 & 98.5\\
				\textbf{LG-SPIL(Ours)} \ &  {97.5}  &  \textbf{95.3} & \textbf{98.8}\\
				\bottomrule 
		\end{tabular}}
	\end{center}
\end{table}

\begin{table}[]
	\begin{center}
		\caption{Comparison with state-of-the-arts on the AVDV dataset.}\label{table8}
		\scalebox{1.0}{
			\begin{tabular}{cc}
				\toprule  
				\toprule  
				Method \  & \ Acc.(\%)\\
				\midrule  
				TSN~\cite{tsn} \  & \ 92.3\\
				I3D~\cite{i3d} \ & \ 93.1\\
				3D-ResNet101~\cite{hara2018can} \  & \ 94.5\\
				ECO~\cite{eco} \  & \ 93.1\\
				Representation Flow~\cite{piergiovanni2019representation} \ & \ 94.8\\
				Flow Gated Network~\cite{rwf} \  & \ 91.1\\
				TEA~\cite{tea}  \ & \ 95.9\\
				TIN~\cite{tin}  \ & \ 94.3\\
				\midrule  
				PointNet++~\cite{qi2017pointnet++} \  & \ 89.7\\
				PointConv~\cite{pointconv} \  & \ 90.2\\ 
				DGCNN~\cite{wang2019dynamic} \  & \ 92.3\\
				L-SPIL~\cite{su2020human} \ & \ {95.1}\\
				\textbf{LG-SPIL(Ours)} \  & \ \textbf{96.2}\\
				\bottomrule 
		\end{tabular}}
	\end{center}
\end{table}

We compare our method with state-of-the-art approaches. Here, we adopt several models that perform well in traditional video action recognition tasks, and apply them to the violent video recognition tasks in this paper (they all are pre-trained on Kinetics~\cite{kay2017kinetics}). At the same time, we also compared the excellent models in processing standard 3D point clouds tasks with our method.

\begin{figure*}
	\begin{center}
		\centering
		\includegraphics[width=6.6in]{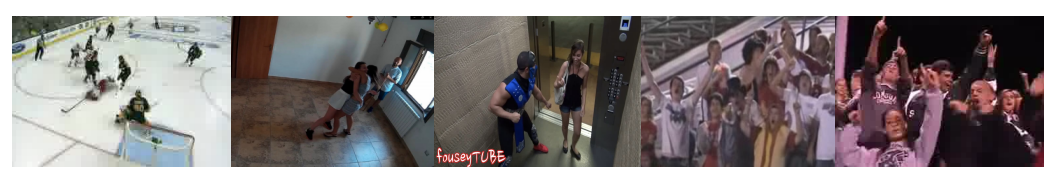}
	\end{center}
	\caption{Some non-violence confusing examples in the datasets. From left to right: “hockey ball scrambling", “hug", “dance", “high five" and “wave" movement that are physically contacted with human bodies.}
	\label{fig8}
\end{figure*}

\begin{figure*}
	\begin{center}
		\centering
		\includegraphics[width=6.8in]{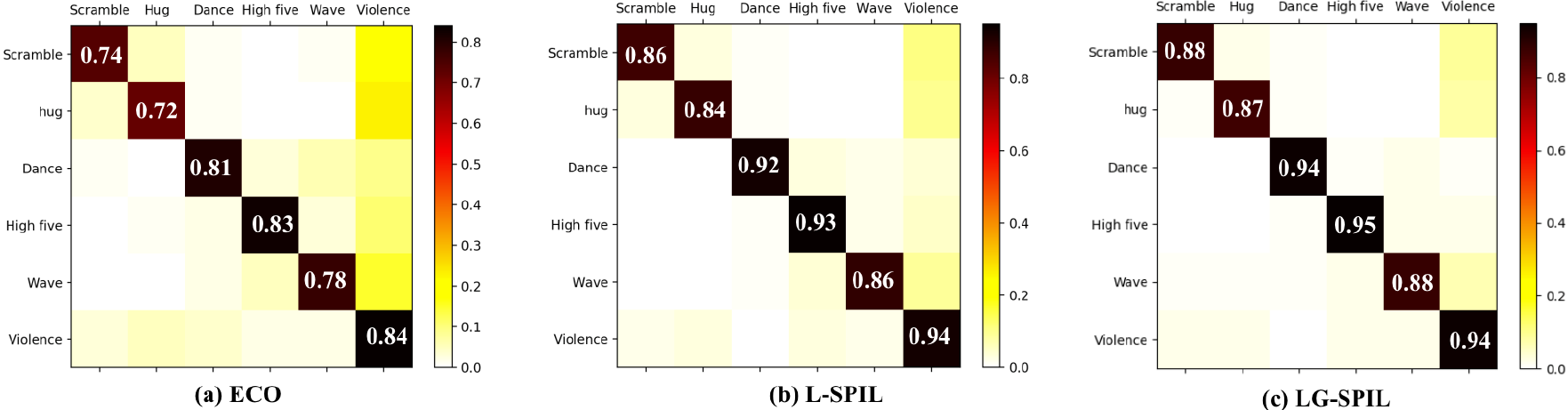}
	\end{center}
	\caption{Confusion matrix comparison on the confusing  samples. (a) ECO~\cite{eco} method. (b) L-SPIL~\cite{su2020human} method. (c) Our LG-SPIL method.}
	\label{fig9}
\end{figure*}

As shown in table~\ref{table6} the experiments on RWF-2000~\cite{rwf} dataset. It can be seen  above the solid line that although these methods reach the state-of-the-art level in general video action recognition tasks, their best results can only reach \textbf{87.3\%} in violent video recognition which is about  3\% lower than our proposed method with accuracy in \textbf{90.0\%}.
This shows that traditional action video recognition methods lack the ability to extract the dynamic characteristics of people in the violence videos and the long-term correlation performance in each frame. 
At the same time, in violent videos, due to the diversity of characters and the variety of scenes, as a result, these methods rely on information such as optical flow~\cite{brox2004high} information, global scene characteristics are invalidated, resulting in low recognition results.
When we use human skeleton points to identify action characteristics, previous point clouds methods are not targeted and are not sensitive to the skeleton points. Therefore, the recognition results are not accurate enough. By learning different skeleton points interactions through our method, we can reach the leading level compared to the graphs NNs, MLPs, or heatmaps-based CNNs methods.

We further evaluate the proposed model on the Hockey-Fight~\cite{hockey}, the Crowd Violence~\cite{crowd} and the Movies-Fight~\cite{movie} dataset.
As shown in table~\ref{table7}, specifically, in terms of accuracy, the average level on the crowd dataset is lower than the other Hockey-Fight and Movies-Fight datasets. This is because the crowd dataset has more people and its background information is relatively complicated. Therefore, some previous work is difficult to have a high degree of recognition. Our LG-SPIL module is not affected by complex scenes. It obtains features by extracting the skeleton point clouds information of the people, which can achieve better performance. 

Finally, we conduct experiments on the AVDV dataset~\cite{bianculli2020dataset}, which is presented for the prevention of false positives. It aims to understand the effectiveness of the violence detection techniques in clips showing rapid moves (hugs, claps, high-fives, etc.) that are not violent. As is shown in table~\ref{table8}, our method still outperforms other approaches. 
The outstanding performance shows the effectiveness and generality of the proposed LG-SPIL for capturing the related points information in multiple people scene. Under different datasets and different environmental scenarios, our method does not rely on other prior knowledge and will not overdone in one dataset and causes inadaptability in other scenarios.

\subsection{Network Complexity}

Compared to L-SPIL~\cite{su2020human}, we insert the global-SPIL module after the local-SPIL module to extract the human movment features. Under the setting of batch-size 8 and 2048 sampled points, one training iteration costs 1.48 seconds with 4 Nvidia 2080Ti GPUs. For inference, it takes 0.32 seconds under the same setting. As shown in table~\ref{table9}, with a few more number of parameters and computation complexity, our proposed method can achieve new state-of-the-art performance. 

\begin{table}[]
	\begin{center}
		\caption{The number of parameters and computation complexity of the network.}\label{table9}
		\scalebox{0.95}{
			\begin{tabular}{ccc}
				\toprule  
				\toprule  
				Method  & Parameter (M)  & GFLOPs   \\
				\midrule  
				L-SPIL~\cite{su2020human} & 25.54  & 33.68  \\
				LG-SPIL (Ours) &   27.18 & 35.26  \\
				\bottomrule 
		\end{tabular}}
	\end{center}
\end{table}

\subsection{Visualization}

\begin{figure}
	\begin{center}
		\centering
		\includegraphics[width=3.3in]{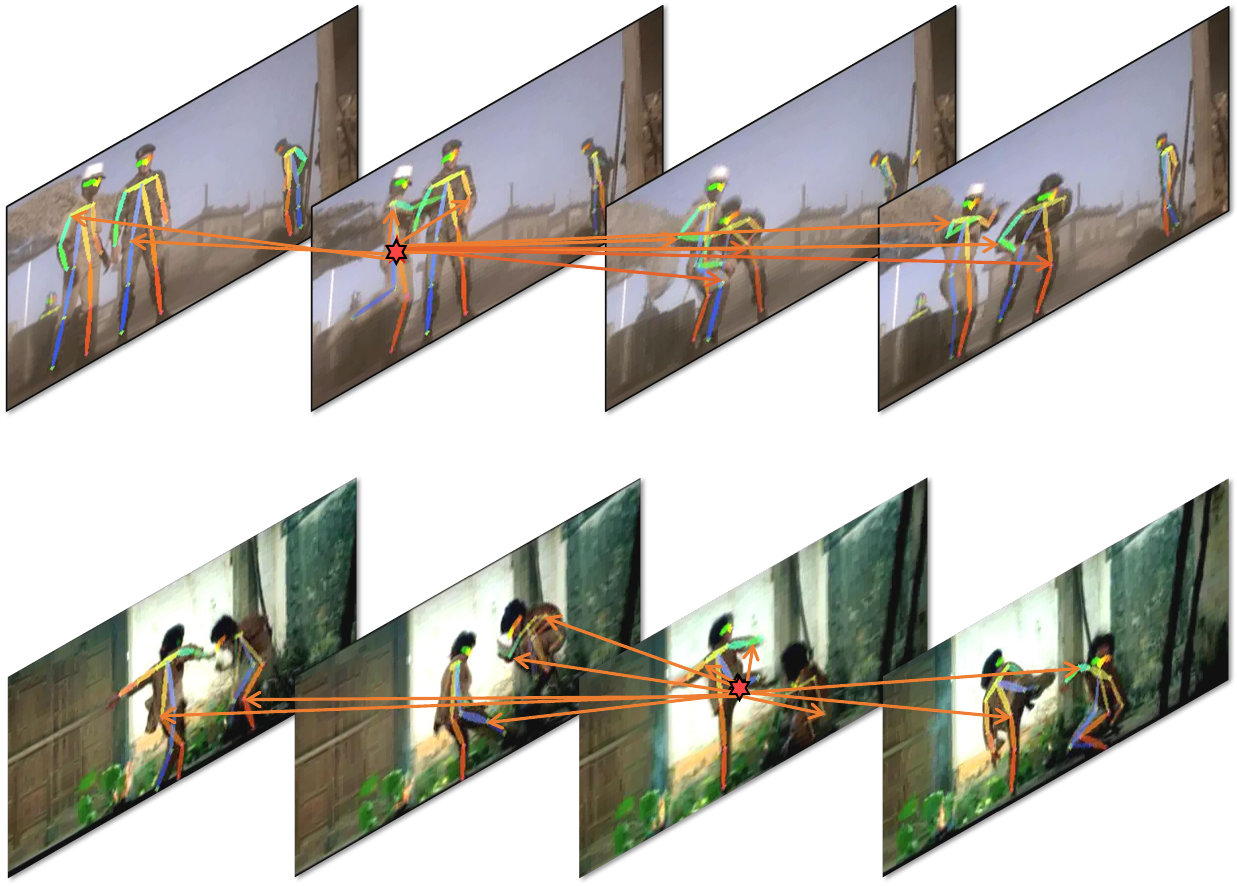}
	\end{center}
	\caption{Examples of skeleton point clouds interaction learning computed by the 2$^{nd}$ layer of Local-SPIL modules. The starting point of arrows represents one sampled centroid point. The top-10 highest weighted arrows for each centroid point are visualized.}
	\label{fig10}
\end{figure}

In Fig~\ref{fig10}, we provide some visualization examples of the learning process. The red star represents the sampled centroid point of the human skeleton. The arrows represent the top-$10$ potentially corresponding skeleton points. As we can see, our algorithm can establish semantic correspondence between skeleton points, thus contributing more to the movements. These visualizations show how the model finds related clues to support its recognition.

\subsection{Failure Case}

To further analyze the network, we study the misclassification results of the above datasets.
We found that some similar to violent actions are misclassified as violence. As shown in Fig~\ref{fig8}, some movements in the images that have physical contact, the network may consider they are fighting or some other violence.
Therefore, we collect all the samples from five datasets evenly and formed them into a new dataset for recognition.

Fig~\ref{fig9} shows the confusion matrix comparison. In those actions that are very similar to violence but are not actual acts of violence, compared with the traditional video action recognition technique, our method can have a large degree of discrimination to classify these actions.
The above results illustrate that the proposed network is an effective method for video violence recognition.
Essentially paying more attention to the information about the skeletal modality movements of the characters will help to distinguish the behaviors of the characters.

\subsection{More-In-Depth Discussion}

Our proposed method is designed for solving the problems in video violence recognition task which is quite different from standard action recognition in two aspects: 1) Multi-human interaction. 2) Complex scenarios.
Therefore, we proposed two types of Skeleton Points Interaction Learning (SPIL) strategies to extract the human features for video violence recognition.
These two modules aim to model the \textbf{interaction} between each person in the scenarios and then extract the feature of them to conduct violent action recognition.
Compared to the standard action recognition dataset, which only has a single person in the simple scenarios in most cases, violent-related datasets always contain multi-people in the scenarios with complicated backgrounds. Therefore, our proposed Skeleton Points Interaction Learning (SPIL) strategies can capture the long-term relationships between each skeletal point. Note that our contributions lie in modeling the interaction of human, standard datasets are consist of most the individual actions, thus, when testing on non violent-related dataset, it can not show the advantages of our method.

\section{Conclusion}
In this paper, we propose a novel and effective approach for video violence recognition. To the best of our knowledge, we are the first one to solve this task by using a 3D point clouds technique to extract action feature information of human skeleton points. 
By introducing the Skeleton Points Interaction Learning (SPIL) module, our model is able to assign different weights according to different skeleton points to obtain the motion characteristics of different people.
Furthermore, we also design a multi-head mechanism to process different types of information in parallel and eventually aggregate them together.
The experiment results on four violent video datasets are promising, demonstrating the effectiveness of our proposed method and our network outperforms the existing state-of-the-art violent video recognition approaches.

\ifCLASSOPTIONcaptionsoff
  \newpage
\fi

\bibliographystyle{IEEEtran}
\bibliography{mybib}

\end{document}